\pdfoutput=1
\documentclass[]{article}
\usepackage[letterpaper]{geometry}
\usepackage{mtsummit2021}
\usepackage{times}
\usepackage{url}
\usepackage{latexsym}
\usepackage{natbib}
\usepackage{layout}
\usepackage[dvipsnames]{xcolor}
\usepackage{todonotes}
\usepackage{haldefs}
\usepackage{booktabs}
\usepackage{subcaption}
\usepackage{tabularx}
\usepackage{hyperref}
\usepackage{wrapfig}
\usepackage{fancyhdr}

\newcommand{\newpara}[1]{\noindent \textbf{#1} \hspace{0.5em}}

\renewcommand{\sectionautorefname}{\S\kern-0.2em}
\renewcommand{\subsectionautorefname}{\S\kern-0.2em}
\renewcommand{\subsubsectionautorefname}{\S\kern-0.2em}

\usepackage{microtype}

\begin{document}

\title{Structural Biases for Improving Transformers on Translation into Morphologically Rich Languages}
\author{\name{\bf Paul Soulos\thanks{\:\:Work partially done while at Microsoft Research.}$^{\ \ 1}$},
        \name{\bf Sudha Rao$^{2}$},
       \name{\bf Caitlin Smith\footnotemark[1]$^{\ \ 1}$},
    \name{\bf Eric Rosen\footnotemark[1]$^{\ \ 1}$},
    \name{\bf Asli Celikyilmaz$^{2}$},
    \name{\bf R. Thomas McCoy\footnotemark[1]$^{\ \ 1}$},
    \name{\bf Yichen Jiang\footnotemark[1]$^{\ \ 3}$},
    \name{\bf Coleman Haley\footnotemark[1]$^{\ \ 1}$},
    \name{\bf Roland Fernandez$^{2}$},
    \name{\bf Hamid Palangi$^{2}$},
    \name{\bf Jianfeng Gao$^{2}$},
    \name{\bf Paul Smolensky$^{1,2}$}\\
    $^{1}$Johns Hopkins University~~~~~$^{2}$Microsoft Research, Redmond~~~~~$^{3}$UNC Chapel Hill \\
 \small\texttt{\{psoulos1, csmit372, erosen27, tom.mccoy, chaley7\}@jhu.edu} \\
 \small\texttt{\{sudha.rao, aslicel, rfernand, hpalangi, jfgao, psmo\}@microsoft.com}\\
 \small\texttt{\{yichenj\}@cs.unc.edu}
}

\maketitle
\pagestyle{empty}

\pagestyle{fancy}
\fancyhf{}
\fancyhead{}
\newcommand{\changefont}{%
    \fontsize{6}{6}\selectfont
}
\fancyfoot[C]{\changefont \textit{Proceedings of the 18th Biennial Machine Translation Summit, Virtual USA, August 16 - 20, 2021\\
4th Workshop on Technologies for MT of Low Resource Languages}}
\renewcommand{\headrulewidth}{0pt}

\begin{abstract}
Machine translation has seen rapid progress with the advent of Transformer-based models. These models have no explicit linguistic structure built into them, yet they may still implicitly learn structured relationships by attending to relevant tokens. We hypothesize that this structural learning could be made more robust by explicitly endowing Transformers with a structural bias, and we investigate two methods for building in such a bias. One method, the TP-Transformer, augments the traditional Transformer architecture to include an additional component to represent structure. The second method imbues structure at the data level by segmenting the data with morphological tokenization. We test these methods on translating from English into morphologically rich languages, Turkish and Inuktitut, and consider both automatic metrics and human evaluations. We find that each of these two approaches allows the network to achieve better performance, but this improvement is dependent on the size of the dataset. In sum, structural encoding methods make Transformers more sample-efficient, enabling them to perform better from smaller amounts of data.
\end{abstract}

\section{Introduction}

The task of machine translation has seen major progress in recent times with the advent of large-scale Transformer-based models \cite[e.g.,][]{Vaswani2017AttentionIA,dehghani2018universal,liu_deep_2020}. However, there has been less progress on language pairs that specifically involve morphologically rich languages. Moreover, although there has been previous work that builds linguistic structure into translation models to deal with morphological complexity \citep{sennrich-haddow-2016-linguistic,dalvi-etal-2017-understanding,matthews-etal-2018-using}, to the best to our knowledge there has not been work that applies such strategies to large-scale Transformer-based models. We hypothesize that providing Transformers access to structured linguistic representations can significantly boost their performance on translation into languages with complex morphology that encodes linguistic structure.

In this work, we investigate two methods for introducing such structural bias into Transformer-based models. In the first method, we use the TP-Transformer (TPT) \citep{schlag2019enhancing}, in which a traditional Transformer is augmented with Tensor Product Representations (TPRs) \citep{smolensky1990tensor} (\autoref{sec:tp-transformer}).
At a high level, TPRs use a composition of \textit{roles} and \textit{fillers} where roles encode structural information (e.g., the part-of-speech of a word) and fillers encode the content (e.g., the meaning of a word). This enables learned internal structured representations. In the second method, we encode structure external to the model by segmenting training data using morphological tokenization (\autoref{sec:morphological-segmenter}): morphological segmentation is done by existing parsers prior to training the Transformer. Since all neural models that operate over sequences tokenize the training data, through this method, we aim to answer the question of whether linguistically-informed tokenization that respects morphological structure can be helpful in processing morphologically-rich languages. 
Through the use of TPT, we aim to examine whether enabling a Transformer to learn its own structured \textit{internal} representations will help it learn linguistic structure including structure which is encoded morphologically in morphologically-rich languages. 
Unlike the morphological tokenizer, the TPT architecture is language-agnostic and can be used on arbitrary datasets without feature engineering. We further investigate how the biases of these two approaches work together.
We experiment on the task of translating from English into two morphologically rich languages: Turkish and Inukitut (Inuit; Eastern Canada). For Turkish, we train on several different dataset sizes from Open Subtitles (1.4M, 5M and 36M), a spoken-language domain, and also fine-tune on SETimes (200K), a news-wire domain. For Inuktitut, we train on the Nunavut Hansard Corpus (1.3M). We test models' performance using both an automatic metric and human evaluation (\autoref{sec:experimental-results}). 

\begin{table}[]
\footnotesize
    \centering
    \begin{tabularx}{\columnwidth}{l X}
    \toprule
     English: & I want people to raise their hands who are in favour of the motion to report progress. (17) \\
     Turkish: & Ilerleme raporunun talep edilmesinden yana olanların el kaldırmalarını istiyorum. (9) \\
     Inuktitut: & isaaquvaksi taikkua nangmaksaqtut pigiaqtitausimajumut nuqqarumaliqtu. (5) \\
     \bottomrule
    \end{tabularx}
    \caption{Parallel sentence in English, Turkish, and Inuktitut. The number of words in each translation (marked in parentheses) is indicative of their information density and, hence, their morphological complexity.}
    \label{tab:translation-example}
\vspace{-1em}
\end{table}

In the English to Turkish translation task, we find that the TP-Transformer beats the Transformer when evaluated for nuances such as morphology, word-order and subject/object-verb agreement. TPT provides a significant improvement on small datasets segmented with language agnostic BPE ($\sim$~1 BLEU for Open Subtitles 1.4m and $\sim$~2.5 BLEU for Hansard) and a more modest improvement on larger datasets (0.16 BLEU for Open Subtitles 5m and 0.36 BLEU for Open Subtitles 36m). Using morphologically segmented data helps substantially with models that are trained on small datasets. This is true for both pre-training (Open Subtitles 1.4m and Inuktitut Hansard), as well as models that are trained on large datasets and later finetuned using a smaller dataset (SETimes). This suggests that the method of encoding structure directly in the training data helps substantially with sample efficiency and transfer learning.

In order to better understand our models, we conduct detailed analysis, including error analysis, on sample outputs from different model variations (Appendix \ref{sec:output-analysis}). We also separate results out into different bins as defined by the morphological density of the target outputs to understand how results vary with morphological complexity \autoref{sec:morph-density}. We find that morphological tokenization is strongly correlated with improved performance on complex sentences.
\section{Using the TP-Transformer}\label{sec:tp-transformer}

The TP-Transformer (TPT) 
was introduced by \citet{schlag2019enhancing} to improve performance on mathematical problem solving, a highly symbolic task. The model introduces an additional component to the attention mechanism which represents relational structure. In addition to the standard key $K$, query $Q$, and value $V$ vectors used in attention, they introduce the role vector $R$. Let the input for token $i \in {1,..,N}$ at layer $l$ be represented as $X_i^{l}$. For head $h$, the vectors are:
\begin{align*}
Q_i^{lh} = X_i^{l}W_q^{lh} + b_q^{lh} \hspace{5mm}
K_i^{lh} = X_i^{l}W_k^{lh} + b_k^{lh} \hspace{5mm}
V_i^{lh} = X_i^{l}W_v^{lh} + b_v^{lh} \hspace{5mm} 
R_i^{lh} = X_i^{l}W_r^{lh} + b_r^{lh}
\end{align*}

The output of soft attention $\bar{V}_i^{lh}$ is: $\bar{V}_i^{lh} = \sum_{t=1}^N \mathrm{softmax}(\frac{Q_i^{lh}K_t^{lh}}{\sqrt{d_k}})^{\top}V_t^{lh}$

In a Tensor Product Representation, role vectors are bound to their corresponding filler vectors by the tensor product $\otimes$ or some compression of it: in the TPT, we use the compression of discarding the off-diagonal elements, resulting in the elementwise or Hadamard product $\odot$. 
The query $Q_i^{lh}$ is interpreted as probing for a filler for the role $R_i^{lh}$, so the output of attention $\bar{V}_i^{lh}$ is taken to be the filler of that role; thus for the original TPT, this yielded: $Z_i^{lh} = \bar{V}_i^{lh} \odot  R_i^{lh}$.

The role vector $R$ is intended to act as a structural encoding independent of that structure's content (which is encoded in $\bar{V}$). We hypothesize that, by disentangling structure and content in this way, we can improve the model's ability to place familiar linguistic units in novel structures (e.g., using a suffix with a word stem that never had that suffix during training). Such structural flexibility is crucial for morphologically-rich languages. 

\begin{wrapfigure}{l}{0.5\textwidth}
    \begin{center}
    \includegraphics[scale=.26]{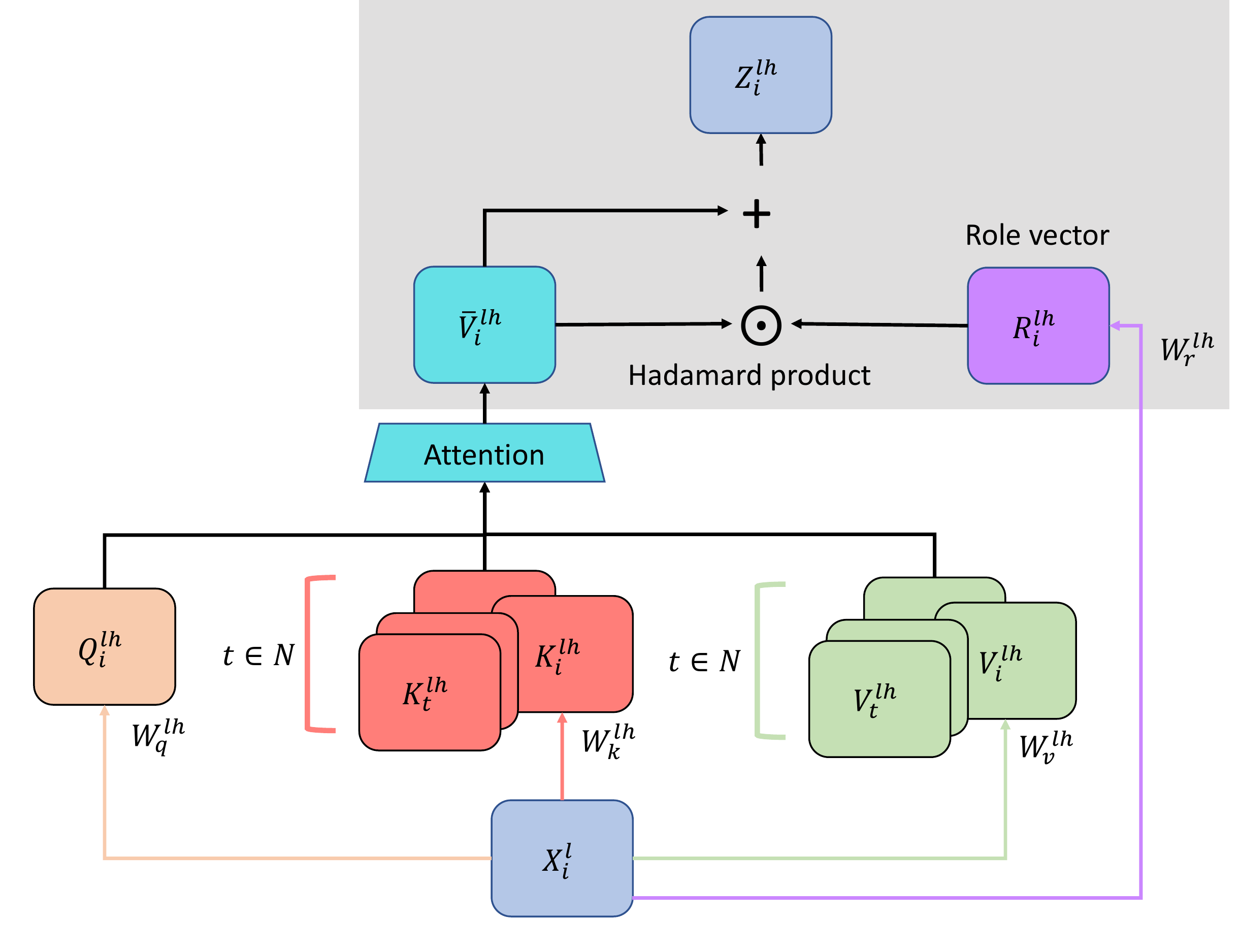}
    \end{center}
    \caption{Architectural diagram of TPT attention mechanism. Highlighted section shows the additional components added to standard Transformer attention.}
    \label{fig:tpt-attention}
    \vspace{-2em}
\end{wrapfigure}

We make two modifications to the TPT used in \citet{schlag2019enhancing}\footnote{Code available at \url{https://github.com/psoulos/tpt}}. First, we use relative position embeddings \citep{shaw-etal-2018-self}. We also use a residual connection to produce gradients that are not zero; $\bar{V} \odot R$ is a multiplicative interaction, so values of $R$ near $0$ will produce activation values and gradients of $0$. This is similar to the model detail in \citet{film} Section 7.2. 
A schematic of our attention is shown in Figure \ref{fig:tpt-attention}. The rest of the architecture follows the standard residual connections and encoder-decoder architecture defined in \citet{Vaswani2017AttentionIA}

\section{Using morphological segmentation}\label{sec:morphological-segmenter}

\begin{wraptable}{r}{0.5\textwidth}
\vspace{-.5em}
\footnotesize
\resizebox{0.5\textwidth}{!}{
\begin{tabular}{ |ll| } 
 \hline
 \textbf{Language} & \textbf{Segmented Word} \\ 
 \hline
 Turkish & anla-t-ma-yacak \\
 Gloss & understand-\textsc{Caus}-\textsc{Neg}-\textsc{Fut} \\
 English & will not tell \\ 
 BPE & anlat-mayacak \\
 \hline
 Inuktitut & miv-vi-liar-uma-lauq-tur-uuq \\
 Gloss & land-place-go-want-\textsc{Past}-3\textsc{s}-say.3\textsc{s} \\
 English & He said he wanted to go to the landing strip. \\ 
 BPE & mivvi-lia-ruma-lau-qturuuq \\
 \hline
\end{tabular}
}
\caption{Morpheme breakdown, gloss, English, and BPE tokenization of Turkish and Inuktitut morphologically complex words}\label{tab:morphemes}
\vspace{-1em}
\end{wraptable}

Our target languages, Turkish and Inuktitut, both exhibit a high degree of morphological complexity. Words in both languages consist of a root followed by potentially many suffixes, each of which may have multiple surface forms.


We used two methods of subword tokenization: one utilizing a type of character-level byte-pair encoding \citep{bpe}, and one incorporating morphological parsing plus byte-pair encoding. The first method (which we label 'BPE') used SentencePiece \citep{Kudo2018}, a tokenizer that builds subword tokens using a combination of byte-pair encoding and unigram language modeling. BPE relies only on character frequencies and incorporates no morphological information.

The second method (which we call `morphological tokenization') incorporated morphological information by parsing all words (i.e. breaking them up into their composite morphemes) in our morphologically complex target languages before tokenizing them. For Turkish, we used the morphological parser from Zemberek \citep{Akn2007}, an open-source Turkish NLP toolkit. Zemberek uses sentence-level disambiguation to produce the most likely parse of each word given its sentential context. For Inuktitut, we used the morphological parsing method adopted by \citet{joanis-etal-2020-nunavut}, incorporating a symbolic parser with a neural parser backoff. See Appendix \ref{sec:morph-process} for implementational details on morphological segmentation.

The differences in how these tokenizers divide multi-morphemic Turkish and Inuktitut words into subwords are illustrated in Table \ref{tab:morphemes}. The boundaries determined by BPE do not reflect the internal morphological structure of these words.

\section{Dataset description}

\subsection{English-Turkish data}

For pretraining of the English-Turkish translation model, we used the Open Subtitles corpus \citep{lison-tiedemann-2016-opensubtitles2016}. This corpus consists of a large number of aligned pairs of subtitles from film and television. In order to test the effect of dataset size on model performance, we utilized three splits of this corpus: the full-size corpus, a sample of five million sentence pairs, and a sample of approximately one million sentence pairs. For fine-tuning of the English-Turkish model, we used the South-East European Parallel (SETimes) Corpus. SETimes is a collection of short written news stories in ten languages. For this task, we used the subset of this corpus that was used for the WMT 2018 English-Turkish shared translation task \citep{bojar-EtAl:2018:WMT1}.

\subsection{English-Inuktitut data} \label{sec:inuk-data}

\begin{wraptable}{l}{0.5\textwidth}
\resizebox{0.5\textwidth}{!}{
\begin{tabular}{ |cccc| } 
 \hline
 Corpus & Training & Validation & Test \\ 
 \hline
 Open Subtitles 36m & 28,694,211 & 3,586,776 & 3,586,777 \\ 
 Open Subtitles 5m & 4,000,000 & 500,000 & 500,000 \\ 
 Open Subtitles 1.4m & 1,300,000 & 65,000 & 65,000 \\ 
 SETimes & 207,678 & 3,007 & 3,000 \\
 Nunavut Hansard & 1,312,791 & 5,494 & 6,181 \\ 
 \hline
\end{tabular}
}
\caption{Number of training, validation, and test samples in the different datasets.}\label{tab:dataset-size}
\vspace{-1em}
\end{wraptable}

Like Turkish, Inuktitut is a morphologically complex language. Words may consist of a root, a prefix, and potentially many suffixes. \autoref{tab:morphemes} contains an example of a multi-morphemic Inuktitut word. For training of the English-Inuktitut translation model, we used the Nunavut Hansard Inuktitut–English Parallel Corpus 3.0 \citep{joanis-etal-2020-nunavut}, the only sizable publicly available bilingual corpus. The dataset consists of over one million aligned sentence pairs from government proceedings. The size of the dataset splits  are reported in \autoref{tab:dataset-size}.

\section{Experimental Results}\label{sec:experimental-results}

We aim to answer the following research questions (RQ) through our experimentation:
\begin{enumerate}[noitemsep,nolistsep]
\item Do either or both of our structural methods improve translation?
\item If so, how does that advantage interact with:
    \begin{enumerate}[noitemsep,nolistsep]
    \item Training data quantity?
    \item Transfer learning?
    \item Morphological richness of language?
    \end{enumerate}
\end{enumerate}

As a baseline, we trained the standard Transformer model \citep{Vaswani2017AttentionIA} with the addition of relative position embeddings \citep{shaw-etal-2018-self}. Model training details and computing resources can be found in Section 1 and 2 of the supplementary materials. For each model, we used either byte pair encoding (BPE) \citep{sennrich-etal-2016-neural} or morphological tokenization as described in \autoref{sec:morphological-segmenter}. In order to see how our changes relate to sample efficiency, we vary the size of the subset of the Open Subtitles dataset used for training. We used the SETimes dataset to finetune these models to test whether either structural bias improves transfer learning. We also trained models on the Inuktitut dataset to compare the results from languages with differing morphological richness.

\subsection{Automatic Metric Results}\label{sec:auto-results}

\begin{wraptable}{l}{0.5\textwidth}
\centering
\footnotesize
\resizebox{0.5\textwidth}{!}{
\begin{tabular}{|ccc|}
\hline
& \textbf{Transformer} & \textbf{TP-Transformer}\\
\hline
1.4m & 7.5 $\pm$ .43 & 8.44 $\pm$ .25 \\
1.4m morph & 16.63 $\pm$ .19 & 16.89 $\pm$ .07 \\
\hline
5m & 18.70 & 18.86 \\
5m morph & 18.84 & 19.19 \\
\hline
36m & 20.95 & 21.31 \\ 
36m morph & 21.05 & 21.32 \\ 
\hline
\end{tabular}
}
\caption{BLEU scores on the test set of Open Subtitles separated by training set size and tokenization method. For the 1.4m runs, we show the mean and standard deviation of three randomly initialized models. The larger datasets only have one run each due to computational resource reasons.}\label{tab:opensub}
\end{wraptable}

\autoref{tab:opensub} shows the test set BLEU\footnote{We calculated BLEU using SacreBLEU \citep{post-2018-call} and the signature is "BLEU+case.mixed+numrefs.1+smooth.exp+tok.13a+version.1.5.0". All models were also tested with CHRF \citep{popovic-2015-chrf} and the results can be found in \autoref{sec:chrf-results}.} scores for the different size splits of the Open Subtitles dataset (Research Question RQ2a). For the smallest data split of 1.4m samples, TPT provides almost 1 BLEU improvement over a standard Transformer. Using a morphological tokenization provides an 8 BLEU improvement on the small split. Using TPT with morphologically tokenized data does not provide any additional benefit on the 1.4m split. For the two larger splits, TPT (across columns) and morphological parsing (across rows) provides minor improvements (0.1--0.36 BLEU), and this improvement becomes more modest when both are combined (top left cell to bottom right cell) (0.49 BLEU on the 5m split and 0.37 on the full 36m split). Next, in order to analyze whether either structural bias helps with transfer learning (RQ2b), we take the best performing models shown in \autoref{tab:opensub} and finetune them on the SETimes dataset.

\begin{wraptable}{l}{0.5\textwidth}
\centering
\footnotesize
\resizebox{0.5\textwidth}{!}{
\begin{tabular}{|ccc|}
\hline
& \textbf{Transformer} & \textbf{TP-Transformer}\\
\hline
5m & 14.19 & 14.25 \\
5m morph & 15.16 & 15.39 \\
\hline
36m & 16.77 & 17.01 \\ 
36m morph & 18.35 & 18.82 \\ 
\hline
\end{tabular}
}
\caption{BLEU scores on the test set of SETimes from models pretrained on OpenSubtitles (5m) and finetuned on SETimes (200K) divided by training set size and tokenization.}\label{tab:setimes}
\end{wraptable}

The BLEU scores for these finetuned models can be seen in \autoref{tab:setimes}. There is a large increase in BLEU score across rows between models that use either BPE encoding or morphological tokenization. This provides further evidence for the findings from the 1.4m split in \autoref{tab:opensub} that morphological tokenization provides a large improvement in low data regimes. While morphological tokenization does not provide much of an improvement during large-scale pretraining, it is beneficial for transfer learning on a smaller domain.

\begin{wraptable}{l}{0.5\textwidth}
\vspace{-.75em}
\centering
\footnotesize
\resizebox{0.5\textwidth}{!}{
\begin{tabular}{|ccc|}
\hline
& \textbf{Transformer} & \textbf{TP-Transformer}\\
\hline
BPE & 14.51 $\pm$ .69 & 15.44 $\pm$ .21 \\
Morphological & 17.26 $\pm$ .48 & 18.60 $\pm$ .26 \\

\hline
\end{tabular}
}
\caption{BLEU scores on the test set of Inuktitut divided by tokenization. We show the mean and standard deviation of three randomly initialized models.}\label{tab:inuk}
\vspace{-1em}
\end{wraptable}

In addition to Turkish, we trained models on the Inuktitut dataset described in \autoref{sec:inuk-data} to understand the variance of model performance by the morphological richness of languages (RQ2c). We trained models using both data tokenized by BPE encoding as well as by an Inuktitut morphological parser. The results are shown in \autoref{tab:inuk}. As we saw on both the 1.4m Open Subtitles split and SETimes, morphological tokenization provides a substantial improvement in BLEU. TPT provides a large average improvement regardless of the tokenization scheme. Inuktitut is more morphologically complex than Turkish across several measures of morphological complexity\footnote{Using the parallel test sets from \citet{mielke-etal-2019-kind}, we measured a type-token ratio of 0.42 for Inuktitut and 0.19 for Turkish, as well as a relative entropy of word structure of 1.75 for Inuktitut and 1.21 for Turkish} and it is possible that TPT models perform better with more complex morphology. For both Turkish and Inuktitut, TPT models trained on the BPE data score $\sim$1 BLEU higher than standard transformers. When we look at this result for morphological data, we see that the TPT continues to provide a 1 BLEU improvement over a standard Transformer on the more morphologically complex Inuktitut data, but this improvement disappears on the less morphological complex Turkish data.

\subsection{Human-based Evaluation Results}\label{sec:human-results}



 \begin{figure*}[t!]
 \begin{subfigure}{.33\linewidth}
  \centering
     \includegraphics[scale=0.38]{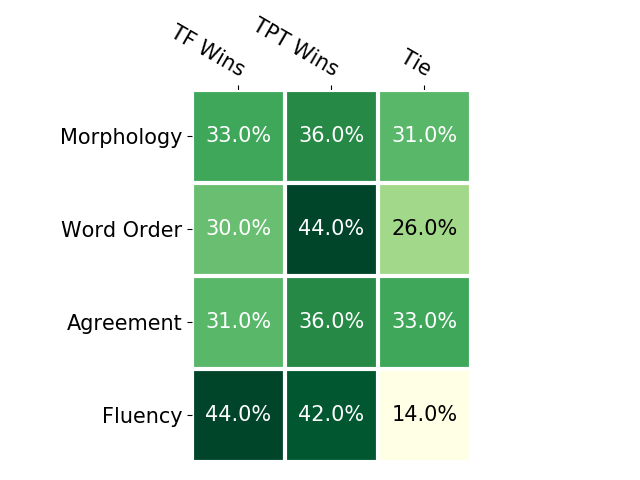}
     \caption{}
     \label{fig:human-eval-5m-BPE}
 \end{subfigure}\hfill
 \begin{subfigure}{.33\linewidth}
  \centering
     \includegraphics[scale=0.38]{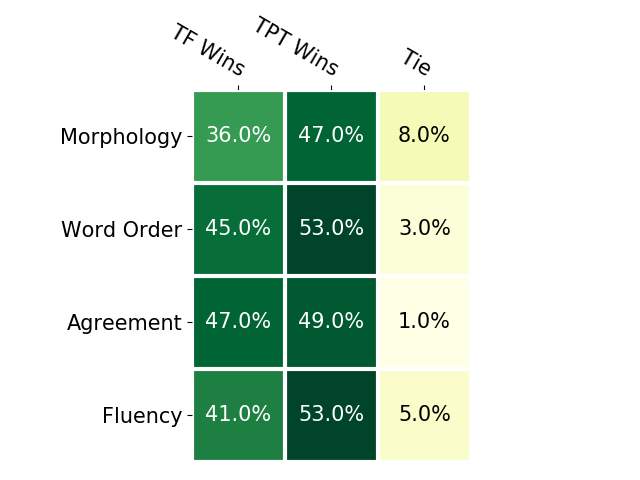}
     \caption{}
     \label{fig:human-eval-5m-MSEG}
 \end{subfigure}\hfill
 \begin{subfigure}{.33\linewidth}
  \centering
     \includegraphics[scale=0.38]{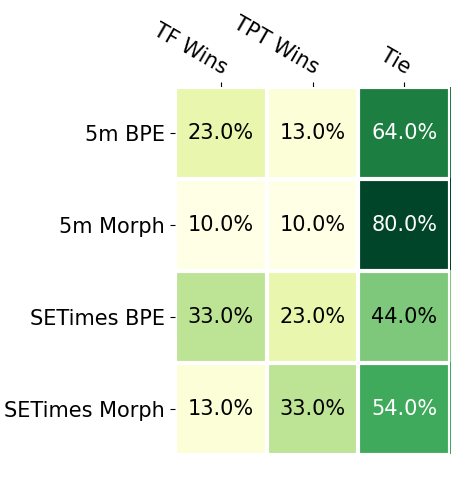}
     \caption{}
     \label{fig:expert-eval}
 \end{subfigure}
 \caption{
 Human judgment results: (a) Comparison between Transformer (TF) and TPT on different criteria when trained on Open Subtitles (5m) using BPE encoding. (b) Comparison between Transformer and TPT when trained on Open Subtitles (5m) using morphologically segmented data. (c) Comparison between Transformer (TF) and TPT on meaning preservation when trained on different datasets. 
 }
 \label{}
 \vspace{-1em}
 \end{figure*}

The BLEU scores in the previous section give us a single number summarizing the quality of our translations. 
We now evaluate some of the finer-grained characteristics of the outputs. We focus on four aspects of the output that are likely to benefit from more robust encodings of structure: morphology, word-order, subject-verb agreement and fluency. 

We use Amazon Mechanical Turk to get human judgements. We perform a comparative study where we show annotators two Turkish
translations from the transformer and the TPT models
trained on the 5m Open Subtitles split. We do not show the English source sentence since the four criteria of evaluation in this study does not require looking at the source sentence. We collect three annotations per comparison and use only those instances where at least 2 out of the 3 annotators agree on the same answer. We collect annotations on 180 instances for each of the two comparative studies. See Appendix \ref{sec:annotator-questions} for the questions asked to annotators.

\autoref{fig:human-eval-5m-BPE} shows the result of this comparison when we use BPE encoding to tokenize the data whereas \autoref{fig:human-eval-5m-MSEG} shows the result of this comparison when we use morphological segmenter to tokenize the data. Under BPE encoding, we find that TPT has slightly less morphological and agreement errors and has significantly less word-order issues. This suggests that the structural bias introduced by the TPT helps in forming sentences that are overall morphologically better formed. On the other hand, annotators find translations from the Transformer to be slightly more fluent than those from the TPT.
Under morphologically segmented data, annotators find translations from TPT are significantly better than the Transformers in  morphological form and word-order and slightly better in subject-verb agreement, providing further evidence that the structural bias introduced by the TPT is helpful. Moreover, annotators also find translations from TPT in this case to be more fluent than those from the Transformer. 

We perform an additional study to understand which of the two model translations best preserves the meaning of the English source sentence.
We ask an expert, a linguistically-trained native Turkish speaker,
to annotate 30 instances each from eight model outputs (5m Open Subtitles BPE \& morphologically tokenized, SETimes BPE \& morphologically tokenized for both Transformer and TPT). 
We show them the English sentence and two Turkish translations. We ask them ``\textit{Grammatical issues aside, which of the two translations better preserves the meaning of the English sentence?}'' and let them choose from A, B or Both preserve equally. 
\autoref{fig:expert-eval} shows the results of this study. In the Open Subtitles dataset, we find the difference between Transformer and TPT performance is too small under both BPE encoding and morphological segmentation. In the SETimes dataset, we find the same trend under BPE encoding. Only under morphologically segmented data in SETimes, TPT significantly wins over Transformer. These results show that when we include the English source sentence, it is inconclusive if TPT or Transformer is better. This suggests that although TPT improves the ability to compose Turkish text (as found by the first study), it does not affect the ability to determine which Turkish output should go with a given English input.

\section{Morphological density analysis} \label{sec:morph-density}

Given the rich morphology of the target languages, we are interested in whether either structural bias or morphological segmentation improves performance on more morphologically complex sentences. To answer this question, we used our Turkish morphological segmenter on sequences from the test set and binned sentences based on the average morphemes per word in a sentence. For example, a long sentence with simple words that are all a single morpheme would have an average morpheme per word of 1, whereas a sentence that is made of complex words would have a larger average morpheme per word. We then calculated the BLEU score for each of these buckets so that we could see if our models performed better on sentences that are morphologically complex.

 \begin{figure*}
      \centering
      \begin{subfigure}[b]{0.2\paperwidth}
          \centering
          \includegraphics[width=0.2\paperwidth]{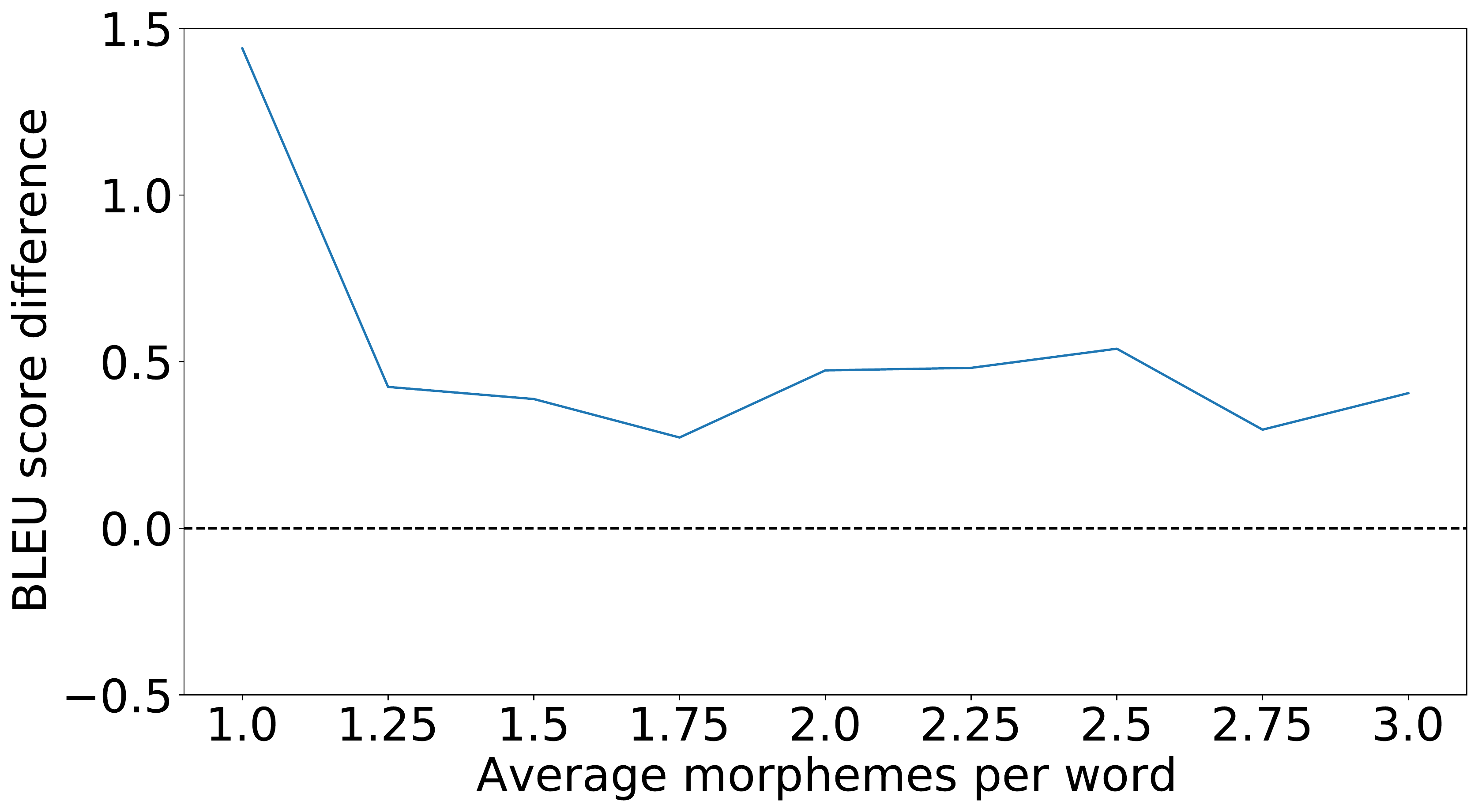}
          \caption{{\tiny 36m TPT BPE $-$ Transformer BPE}}
          \label{fig:32tpt-t5}
      \end{subfigure}
      \begin{subfigure}[b]{0.2\paperwidth}
          \centering
          \includegraphics[width=0.2\paperwidth]{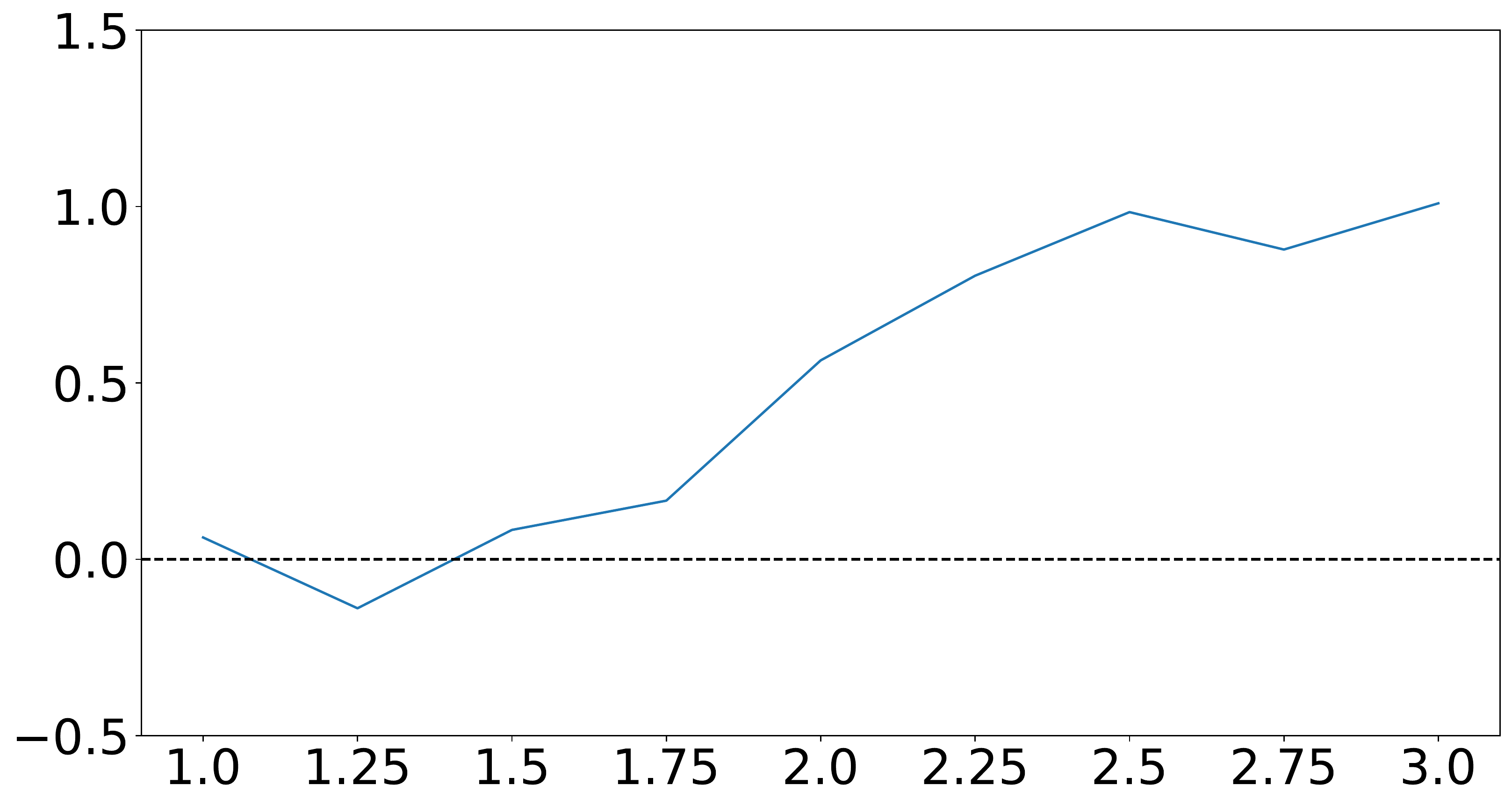}
          \caption{{\tiny 36m Transformer morph $-$ Transformer BPE}}
          \label{fig:32seg-unseg}
      \end{subfigure}
      \begin{subfigure}[b]{0.2\paperwidth}
          \centering
          \includegraphics[width=0.2\paperwidth]{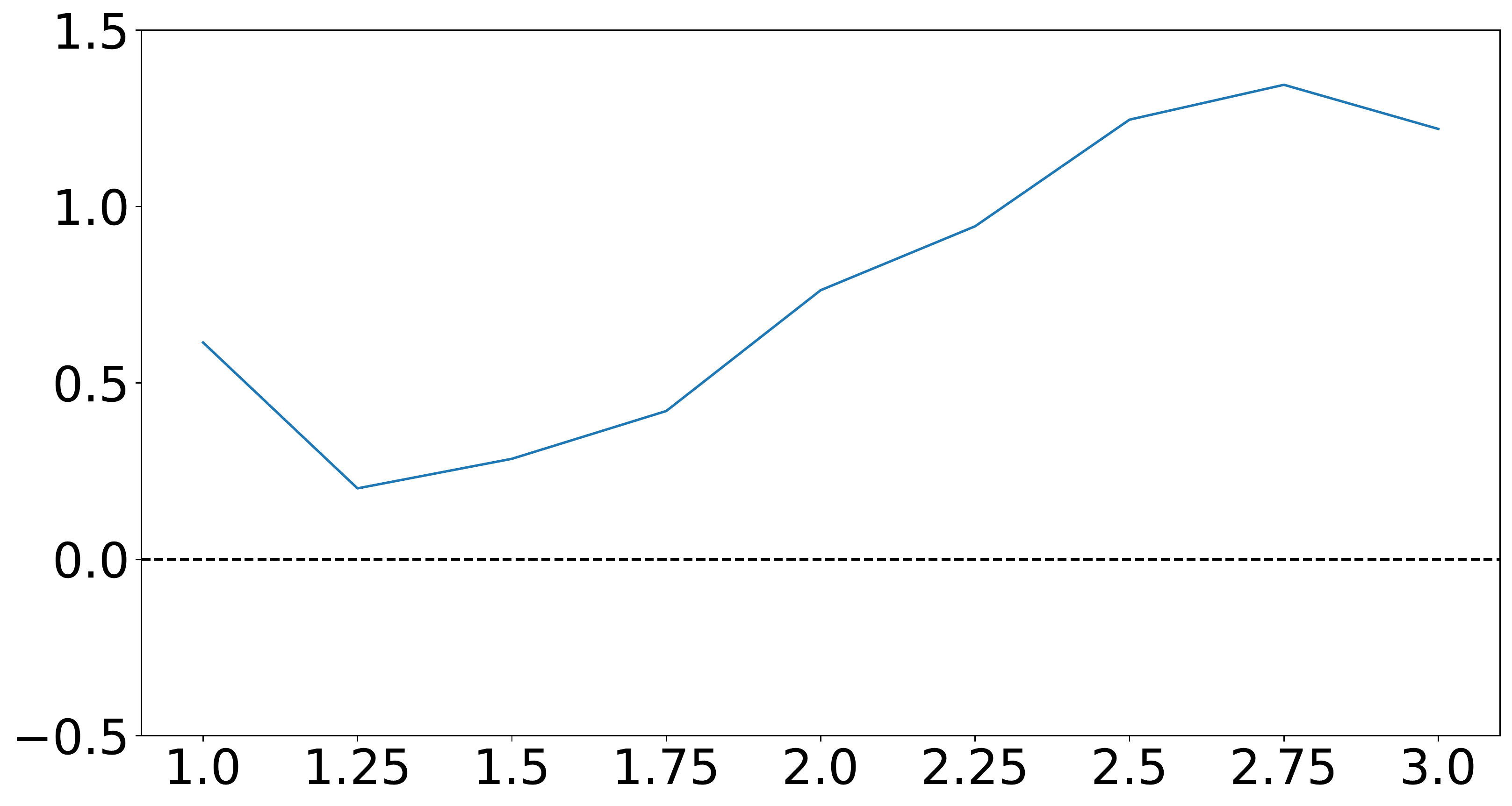}
          \caption{{\tiny 36m TPT morph $-$ Transformer BPE}}
          \label{fig:32tptseg-t5unseg}
      \end{subfigure}
      \begin{subfigure}[b]{0.2\paperwidth}
          \centering
          \includegraphics[width=0.2\paperwidth]{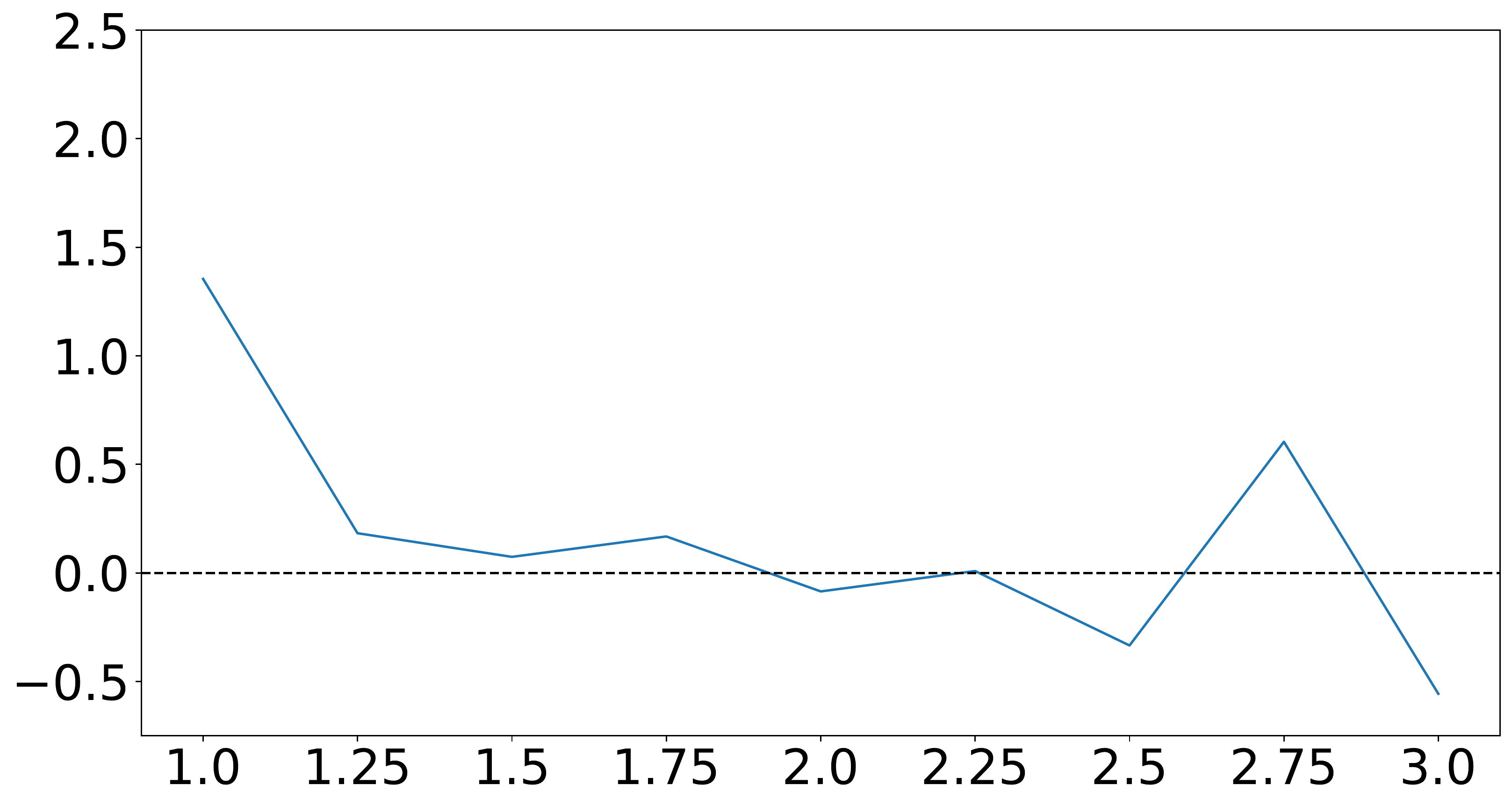}
          \caption{{\tiny 5m TPT BPE $-$ Transformer BPE}}
          \label{fig:5tpt-t5}
      \end{subfigure}
      \begin{subfigure}[b]{0.2\paperwidth}
          \centering
          \includegraphics[width=0.2\paperwidth]{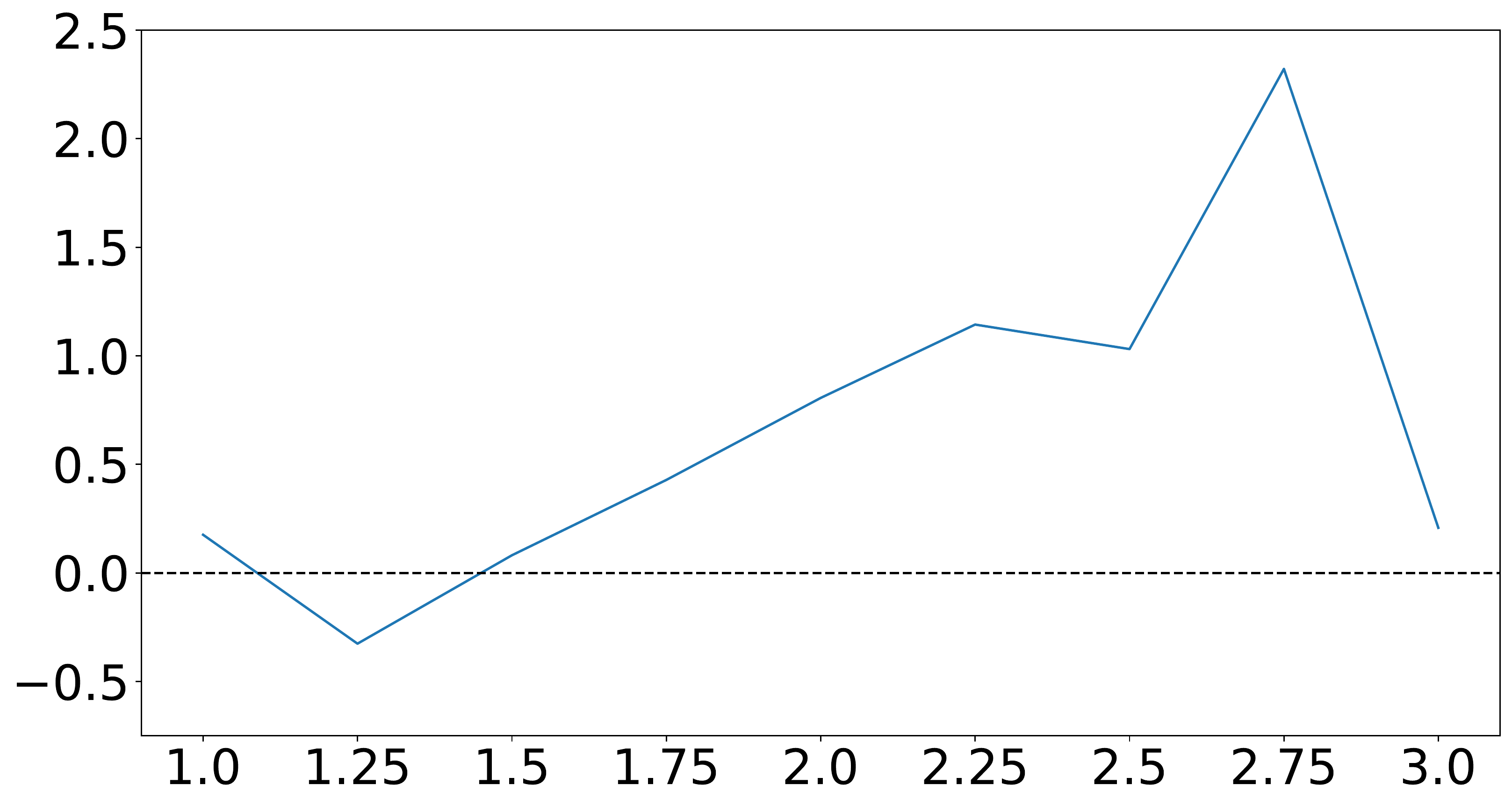}
          \caption{{\tiny 5m Transformer morph $-$ Transformer BPE}}
          \label{fig:5seg-unseg}
      \end{subfigure}
      \begin{subfigure}[b]{0.2\paperwidth}
          \centering
          \includegraphics[width=0.2\paperwidth]{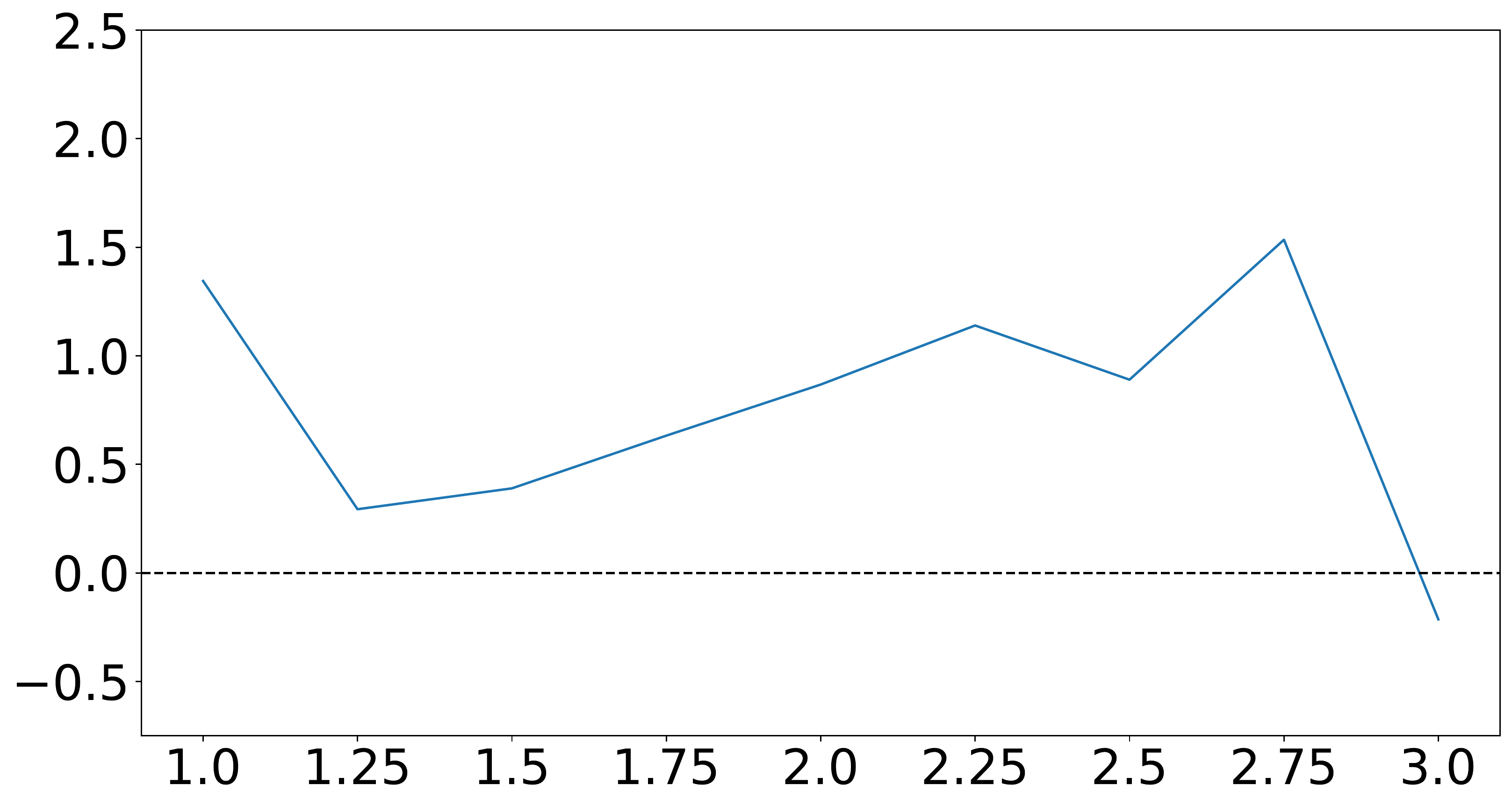}
          \caption{{\tiny 5m TPT morph $-$ Transformer BPE}}
          \label{fig:5tptseg-t5unseg}
      \end{subfigure}
         \caption{BLEU score differences between models on the Turkish Open Subtitles 36m (top row) and 5m (bottom row) training sets bucketed by morphological density (average number of morphemes per word in a sentence).}
         \label{fig:morph-density}
\vspace{-1em}
 \end{figure*}

The results are shown in \autoref{fig:morph-density}. On the 36m training set (top row), both of our methods provide an improvement at almost every morpheme density. Comparing TPT to a standard Transformer, \autoref{fig:32tpt-t5} shows a relatively consistent improvement of around 0.4 BLEU with a large increase for simple sentences. Comparing standard transformers with morphological parsing against BPE, \autoref{fig:32seg-unseg} shows that as the morphological complexity of sequences increases, the model using morphological tokenization improves over BPE tokenization. The same trend is visible when comparing TPT with morphological tokenization with a standard transformer using BPE tokenization (\autoref{fig:32tptseg-t5unseg}), except the magnitude of the increase is greater. 

The morphological analysis on the 5m training set (bottom row)  is less conclusive. TPT does not appear to have any impact as the morphological density increases (\autoref{fig:5tpt-t5}). Morphological tokenization shows a similar upward trend as on the 36m dataset, but this improvement disappears suddenly at 3.0 morphemes per word (\autoref{fig:5seg-unseg}). As the morphological density increases, the number of samples for each bucket on the test set decreases, so it is possible that the sudden drop is the result of too few samples.


Our results also show some correspondence with the overall morphological complexity of the dataset. We computed a modified version of the $C_D$ measure (the ``relative entropy of word structure'') from \citet{bentz-etal-2016-comparison}, as we found it to be the most robust to the meaning variations between corpora (Supplementary materials section 4). Higher values of this measure correspond to more regular structure/information in words, and thus, greater morphological complexity. We computed the measure over the first 100,000 characters of the test set of each dataset. We computed $C_D$ as 1.89 for the Hansard dataset, while the Turkish datasets ranged from $C_D$ 1.45-1.49. This corresponds to the relatively large increase in BLEU seen for Inuktitut.

\section{Related Work}

\paragraph{Translating into Morphologically-rich languages}
Previous work has leveraged morphology for translating into morphologically-rich languages. \citet{turhan} uses a recursive symbolic system to translate from English into Turkish including a morphological generator. \citet{Ataman2020} use hierarchical latent variable models to model both character and morpheme level statistics for translating into morphologically rich languages (Arabic, Czech, Turkish) with GRUs. \citet{passban2018improving} introduce a character-level neural machine translation model for translating into morphologically rich languages which incorporates a morphology lookup table into the decoder whereas \citet{passban2018tailoring} propose a subword-level model that uses separate embedding for stem and affix. \citet{joanis-etal-2020-nunavut} introduced the dataset that we use for Inuktitut and also explored using morphological segmentation for alignment as well as neural and statistical machine translation. This work was followed up by \citet{knowles-etal-2020-nrc} who introduce additional methods techniques on the Inuktitut dataset. \citet{roest-etal-2020-machine} and \citet{scherrer-etal-2020-university} also investigated morphological segmentation in Inuktitut in addition to data augmentation and pretraining.

\paragraph{Using Transformer-based models for translation}

In recent times, there have been several work that use variations of Transformer \citep{Vaswani2017AttentionIA} model for the task of machine translation. \citet{chen2018best} combine the power of recurrent neural network and transformer. 
\citet{dehghani2018universal} introduce universal transformers as a generalization of transformers whereas \citet{deng2018alibaba} combine transformer architecture with several other techniques such as BPE, back
translation, data selection, model ensembling
and reranking. 
\citet{bugliarello2020enhancing} incorporate syntactic knowledge into transformer model to show improvements on English to German, Turkish and Japanese translation tasks. \citet{currey-heafield-2019-incorporating} introduce two methods to incorporate English syntax when translating from English into other languages with Transformers. \citet{liu2020multilingual} introduce mBART, an auto-encoder pretrained on large-scale monolingual corpora and show gains on several languages. 

\paragraph{Using TPRs}

TPRs have gained traction recently with the interest in neurosymbolic computation to achieve out-of-domain generalization. They have been used in a variety of domains, including mathematical problem solving \citep{schlag2019enhancing}, reasoning \citep{schlag2018learning}, image captioning \citep{huang-etal-2018-tensor}, question-answering \citep{Palangi2018QuestionAnsweringWG}, and program synthesis \citep{chen}. A separate line of work uses TPRs as an interpretation tool to understand representations in networks that do not explicitly use TPRs \citep{mccoy2018rnns,soulos-etal-2020-discovering}.

\section{Conclusion}

We investigated two methods for improving translation into morphologically rich languages with Transformers. The TP-Transformer adds an additional component to Transformer attention to represent relational structure. This model had the largest improvement on smaller datasets and modest improvement on larger datasets. We also investigated morphological tokenization which had substantial improvements on small datasets and transfer learning. The models were analyzed by human evaluators to tease apart different dimensions along which our models excel; TP-Transformer had fewer morphological, word-order, and agreement issues. We analyzed the performance of our networks under varying morphological complexity and found that morphological tokenization provides a large benefit for more complex sentences. 

\small

\bibliographystyle{apalike}
\bibliography{mtsummit2021}

\newpage
\appendix
{\Large Appendix}
\section{Model Training Parameters}
Both the standard Transformer and the TP-Transformer (TPT) use 6 layers and 8 heads per layer. TPT has key/value/query/role dimensions of 64, whereas the standard Transformer has key/value/query dimensions of 80. The reason for this increase is so that the resulting models match in terms of parameter count, and we add parameters are the most homologous area. The standard Transformer has 74,375,936 parameters, and the TP-Transformer has 74,385,152 parameters. Both networks use a token dimension of 512, a feedforward dimension of 2048, and 32 relative positioning buckets \cite{shaw-etal-2018-self}. The input vocabulary size is 50,000. We set a training batch size of 80 per GPU and used the Adafactor \cite{adafactor-shazeer18a} optimizer with square root learning rate decay. Throughout the model, we used a commonly used dropout rate of .1.

\section{Computing Resources}
The models were all trained with 8 Tesla V100 GPUs. The models trained on the small Hansard and Open Subtitles 1.4m datasets converged in about 8 hours. The larger Open Subtitles 5m models coverged in around 40 hours, and the Open Subtitles 32m models coverged in 15 days.

\section{Corpora Morphological Complexity}
Studies have considered what corpus-based measures are correlated with linguistic measures of morphological complexity. Most notably, \citet{bentz-etal-2016-comparison} found several corpus-based measures that correlate strongly with complex morphological typology.  This measure computes the regularity of structure within words by taking the character-level entropy of the corpus and subtracting that from the entropy of a ``masked'' version of the corpus, where all non-whitespace characters have been replaced with random samples from the uniform distribution over the characters in the corpus.  Rather than the approximation used in \citet{bentz-etal-2016-comparison} for character-level entropy, we directly computed the character-level Shannon's entropy using a James-Stein shrinkage estimator as in \citet{JMLR:v10:hausser09a}. 

\section{Morphological parser process} \label{sec:morph-process}
For each target language, its parser was used to insert morpheme boundaries into all multi-morphemic words in the dataset. Due to the comparatively low level of morphological complexity of the English source data, no parsing of English words was conducted. From here, each SentencePiece tokenizer's vocabulary was built over a dataset's training data (both the source and target language) with a target size of 50,000 vocabulary items. SentencePiece allows the user to specify special characters that cannot be crossed when constructing subword tokens, both during training of the tokenizer and during tokenization of a sentence. The symbol used to represent morpheme boundaries was specified as such a special symbol. As a result, morpheme boundaries in Turkish and Inuktitut (as identified by their respective parsers) always served as subword token boundaries.

Each SentencePiece tokenizer's vocabulary was built over a dataset's training data (both source and target language) with a target size of 50,000 vocabulary items. This tokenization method (which we label simply `BPE') relies only on character frequencies and incorporates no morphological information, so many multi-morphemic words may each be assigned to a single token, and there is no guarantee that a word's subword boundaries align with its morpheme boundaries.

\section{CHRF Results}\label{sec:chrf-results}
The same models used to measure BLEU scores are also tested using CHRF \citep{popovic-2015-chrf}. The results are shown in Tables \ref{tab:opensub-chrf}, \ref{tab:setimes-chrf}, and \ref{tab:inuk-chrf}.

\begin{wraptable}{r}{0.5\textwidth}
\centering
\footnotesize
\resizebox{0.5\textwidth}{!}{
\begin{tabular}{|ccc|}
\hline
& \textbf{Transformer} & \textbf{TP-Transformer}\\
\hline
1.4m & .351 $\pm$ .005 & .365 $\pm$ .004 \\
1.4m morph & .438 $\pm$ .001 & .440 $\pm$  .001 \\
\hline
5m & .461 & .463 \\
5m morph & .467 & .469 \\
\hline
36m & .486 & .488 \\ 
36m morph & .490 & .492 \\ 
\hline
\end{tabular}
}
\caption{CHRF scores on the test set of Open Subtitles separated by training set size and tokenization method. For the 1.4m runs, we show the mean and standard deviation of three randomly initialized models. The larger datasets only have one run each due to computational resource reasons.}\label{tab:opensub-chrf}
\end{wraptable}

\begin{wraptable}{r}{0.5\textwidth}
\centering
\footnotesize
\resizebox{0.5\textwidth}{!}{
\begin{tabular}{|ccc|}
\hline
& \textbf{Transformer} & \textbf{TP-Transformer}\\
\hline
5m & .502 & .502 \\
5m morph & .509 & .514 \\
\hline
36m & .532 & .537 \\ 
36m morph & .540 & .543 \\ 
\hline
\end{tabular}
}
\caption{CHRF scores on the test set of SETimes from models pretrained on OpenSubtitles (5m) and finetuned on SETimes (200K) divided by training set size and tokenization.}\label{tab:setimes-chrf}
\end{wraptable}

\begin{wraptable}{r}{0.5\textwidth}
\centering
\footnotesize
\resizebox{0.5\textwidth}{!}{
\begin{tabular}{|ccc|}
\hline
& \textbf{Transformer} & \textbf{TP-Transformer}\\
\hline
BPE & .498 $\pm$ .011 & .513 $\pm$ .003 \\
Morphological & .526 $\pm$ .007 & .539 $\pm$ .006 \\
\hline
\end{tabular}
}
\caption{CHRF scores on the test set of Inuktitut divided by tokenization. We show the mean and standard deviation of three randomly initialized models.}\label{tab:inuk-chrf}
\end{wraptable}

\section{Annotator Questions}\label{sec:annotator-questions}
We ask annotators the following questions:

\newpara{Morphology:} ``\textit{Which of the two sentences has more morphological issues (i.e. incorrect suffixes)?}'' and let annotators choose from A, B, Both or None.

\newpara{Word-order:} ``\textit{Which of the two sentences has word-order issues?}'' and let annotators choose from A, B, Both or None.

\newpara{Agreement:} ``\textit{Which of the two sentences has more agreement errors between the subject/object and the verb (i.e. the suffixes for the verbs and/or the nouns do not agree with each other)?}'' and let annotators choose from A, B, Both or None.

\newpara{Fluency:} ``\textit{Which of the two sentences is more fluent i.e. reads more like it was written by a native Turkish speaker?}'' and let annotators choose from A, B, Both are equally fluent.

\section{Output analysis}\label{sec:output-analysis}

\begin{table*}[!ht]
 \centering
 \resizebox{1\textwidth}{!}{%
 \footnotesize
 \begin{tabularx}{\linewidth}{l*{2}{X}}
 \toprule
  &  \textbf{Fluency Issues}\\ \midrule 
  English & \textit{``I want to carry on living," he said at the time of the CPJ award.}\\ 
 Turkish Transformer & \textit{CPJ \textbf{\textcolor{red}{ödülünün zamanında}} konuşan Jovanoviç, ``Yaşamak istiyorum." dedi.}\\
 Turkish TPT & \textit{CPJ ödülünde konuşan bakan, ``Yaşamayı sürdürmek istiyorum." dedi.}\\
 Reason & unnecessary use of the word `zamaninda'.\\
  \midrule \midrule
 &  \textbf{Meaning Preservation}\\ \midrule 
 English & \textit{Some say you chose Turkey for money.}\\ 
 Turkish Transformer & \textit{Bazıları Türkiye'yi \textbf{\textcolor{red}{para karşılığında}} seçtiğinizi söylüyor.}\\
 Turkish TPT & \textit{Bazıları, para için Türkiye'yi seçtiğinizi söylüyorlar.}\\
 Reason & ``para karşılığında" suggests `in exchange for money'\\
 \midrule \midrule
  &  \textbf{Subject to verb agreement Issues}\\ \midrule 
 English & \textit{Maybe because I go to bed listening to the message you left, saying how much you liked missing me.}\\ 
 Turkish Transformer & \textit{Belki de yatağa gidip, beni özlemeyi ne kadar sevdiğini söyleyen mesajını dinlediğim için.}\\
 Turkish TPT & \textit{Belki de yatağa gidip bıraktığın mesajı dinleyip beni özlediğini \textbf{\textcolor{red}{söy-le-di-ğ-im}} için.}\\
 Reason & incorrect use of first person (`-im') instead of second person (`-in')\\
 \midrule \midrule
 &  \textbf{Morphology Issues}\\ \midrule 
  English & \textit{So far we have not received any news nor found any clues.}\\ 
 Turkish Transformer & \textit{Şimdiye kadar hiçbir haber alamadık ve hiçbir ipucu bulamadık}\\
 Turkish TPT & \textit{Bugüne kadar ne haber aldık ne de ipucu \textbf{\textcolor{red}{bul-a-ma-dı-k}}}\\
 Reason & Highlighted word has a double negative instead of the correct form \textbf{\textcolor{ForestGreen}{bul-a-bil-di-k/bul-du-k}}.\\

  \bottomrule
 \end{tabularx}
 }
 \caption{Sample outputs showing issues relating to fluency, meaning preservation, agreement and morphology from Transformer and TPT models.}
  \label{tab:translation-examples}
  \vspace{-1em}
\end{table*}

Here we present an error analysis of a few sample translations from Transformer and TPT models. We group errors according to the aspects used to perform human-based evaluation in \autoref{sec:human-results}. \autoref{tab:translation-examples} shows the result of this analysis. Under fluency issues, Transformer introduces an unnecessary word `zamaninda' making it less fluent compared to the TPT translation. Under meaning preservation, the translation by Transformer incorrectly suggests ``in exchange for money" whereas TPT correctly preserves the meaning. Under agreement issues, TPT includes incorrect use of first person suffix whereas Transformer does not have any subject to verb agreement issues. Under morphology issues, TPT incorrectly includes a negation suffix making the sentence a double negative whereas Transformer correctly translates the English sentence.

\autoref{tab:additional-translation-examples} includes analysis of some additional sample outputs from Transformer and TPT models. Under morphology issues, Transformer includes an unnecessary plural suffix. The TPT translation is okay but would have been better with the addition of the `-mu' suffix. Under meaning preservation, Transformer incorrectly translates ``Bank of England'' as ``Bank of England'', thus losing out on the meaning. Whereas TPT correctly translates that named entity into Turkish. Under tense issues, Transformer uses an incorrect past tense suffix whereas TPT correctly preserves the tense of the English sentence. Under repetition issues, Transformer repeats a word which is not required in written-language but might be okay in spoken-language. 

\begin{table*}[!ht]
 \centering
 \resizebox{1\textwidth}{!}{%
 \footnotesize
 \begin{tabularx}{\linewidth}{l*{2}{X}}
 \toprule
  &  Morphology Issues\\ \midrule 
 \hline
  English & \textit{First we have to decide if those lost six minutes will be coming out of game time, bathroom time or the pizza break.}\\ 
 Turkish Transformer & \textit{İlk önce, bu altı dakika \textcolor{red}{ kaybet-me-ler-i-n} oyun zamanından mı yoksa banyo zamanından mı olacağına karar vermeliyiz.}\\
 Turkish TPT & \textit{İlk olarak, o 6 dakikanın maçtan, banyo saatinden veya pizza molasından \textcolor{green}{(-mı)} çıkıp çıkmayacağına karar vermeliyiz.}\\
 Reason & Unnecessary plural suffix (-ler)  \\
  \hline
  &  Meaning Preservation\\ \midrule 
  \hline
  English & \textit{Bank of England to keep interest rates at 0.25\%}\\ 
 Turkish Transformer & \textit{\textcolor{red}{Bank of England} faiz oranlarını \%0,25 oranında tutacak.}\\
 Turkish TPT & \textit{İngiltere Merkez Bankası faiz oranlarını \%0,25 oranında tutacak.}\\
 Reason & Incorrect translation of named entity\\
 \hline
  &  Tense Issues \\ \midrule 
  \hline
 English & \textit{Barely out of bed and already on the phone.}\\ 
 Turkish Transformer & \textit{Yataktan zar zor çıktım ve \textcolor{red}{telefonla konuştum bile.}}\\
 Turkish TPT & \textit{Yataktan zar zor çıktım ve telefondayım.}\\
 Reason & Incorrect use of past tense suffix (`-tum') instead of present tense suffix (`yorum') \\
 \hline
  &  Repetition Issues \\ \midrule 
   \hline
 English & \textit{Specific criteria, such as an asteroid's size and collision angle, are the factors that would determine the depth of its crater and the damage that its impact would cause.}\\ 
 Turkish Transformer & \textit{Asteroidin büyüklüğü ve çarpışma açısı gibi belli kriterler, kraterin derinliğini \textcolor{orange}{belirleyecek} ve etkisinin yaratacağı hasarı belirleyecek faktörler}\\
 Turkish TPT & \textit{Bir asteroidin büyüklüğü ve çarpışma açısı gibi belirli kriterler, kraterinin derinliğini ve etkisinin yol açacağı hasarı belirleyecek faktörler}\\
 Reason & The word ``belirleyecek'' is repeated which is unnecessary in written-language but would be okay in spoken-language. \\
 \hline
 \hline
 \end{tabularx}
 }
 \caption{Sample outputs  from Transformer and TPT models showing issues relating to morphology, meaning preservation, tense and repetition.}
  \label{tab:additional-translation-examples}
\end{table*}

\end{document}